# Diagnosis of Autism in Children using Facial Analysis and Deep Learning

**Madison Beary, Alex Hadsell, Ryan Messersmith, Mohammad-Parsa Hosseini**

Department of Bioengineering, Santa Clara University—Santa Clara, CA, USA

*Abstract*—In this paper, we introduce a deep learning model to classify children as either healthy or potentially autistic with 94.6% accuracy using Deep Learning. Autistic patients struggle with social skills, repetitive behaviors, and communication, both verbal and nonverbal. Although the disease is considered to be genetic, the highest rates of accurate diagnosis occur when the child is tested on behavioral characteristics and facial features. Patients have a common pattern of distinct facial deformities, allowing researchers to analyze only an image of the child to determine if the child has the disease. While there are other techniques and models used for facial analysis and autism classification on their own, our proposal bridges these two ideas allowing classification in a cheaper, more efficient method. Our deep learning model uses MobileNet and two dense layers in order to perform feature extraction and image classification. The model is trained and tested using 3,014 images, evenly split between children with autism and children without it. 90% of the data is used for training, and 10% is used for testing. Based on our accuracy, we propose that the diagnosis of autism can be done effectively using only a picture. Additionally, there may be other diseases that are similarly diagnosable.

## I. Introduction

**Motivation**: Autism is primarily a genetic disorder, though there are some environmental factors, that causes challenges with social skills, repetitive behaviors, speech, and nonverbal communication. In 2018, the CDC claimed that about 1 in 59 children will be diagnosed with some form of autism. Because there are so many forms of autism, it is technically called autism spectrum disorder (ASD). A child can be diagnosed with ASD as early as 18 months old. Interestingly, while ASD is believed to be a genetic disorder, it is mainly diagnosed through behavioral attributes: "the ways in which children diagnosed with ASD think, learn, and problem-solve can range from highly skilled to severely challenged." Early detection and diagnosis are crucial for any patient with ASD, as this may significantly help them with their disorder.

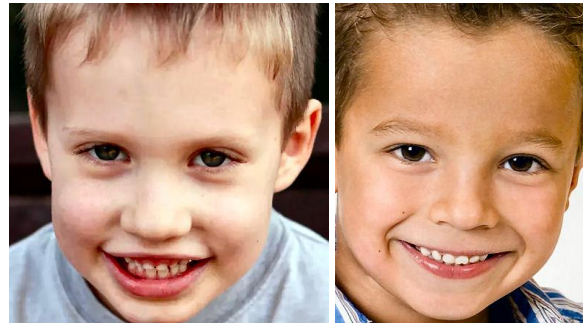

Figure 1: (Left) Child with autism (Right) Child without autism

We believe that facial recognition is the best possible way to diagnose a patient because of their distinct attributes. Scientists at the University of Missouri found that children diagnosed with autism share common facial feature distinctions from children who are not diagnosed with the disease. The study found that children with autism have an unusually broad upper face, including wide-set eyes. They also have a shorter middle region of the face, including the cheeks and nose. Figure 1 shows some of these differences. Because of this, conducting facial recognition binary classification on images of children with autism and children who are labeled as healthy could allow us to diagnose the disease earlier and in a cheaper way.

**Data Information**: For our project, we will be working with a dataset found on Kaggle, which consists of over three thousand images of both autistic and non-autistic children. This dataset is slightly unusual, as the publisher only had access to websites to gather all the images. When downloaded, the data is provided in two ways: split into training, testing, and validation versus consolidated. If we decide to create our own machine learning model, the provided split of training, testing, and validating subgroups will be useful. The validation component will be important for determining the quality of the model we use, which means we do not have to strictly rely on the accuracy of the model to determine its quality. The training, testing, and validating subcategories are further split into autistic and non-autistic folders. The autistic training group consists of 1327 images of facial images, and the non-autistic training directory consists of the same amount of images. The autistic and non-autistic testing directories both have 140 images, for a total of 280 images. Lastly, the validation category has a total of 80 images: 40 non-autistic and 40 autistic facial images. If we can use a model already available, then using the consolidated images would be best, because that will allow us to control the amount used for training and testing.

## II. State of the Art

There have been several studies conducted using neural networks for facial analysis. Most of these studies are focused on determining the age and gender of the individual in question. Additionally, there have been a few studies done focusing on the classification of Autism using brain imaging systems. Our research project has taken the techniques available for facial analysis and applied these to the classification of Autism.

**Facial Analysis**: A study by Wen-Bing Horng and associates worked to classify facial images into one of four categories: babies, young adults, middle-aged adults, and old adults. Their particular study used two back-propagation neural networks for classification. The first focuses on geometric features, while the second focuses on wrinkle features. Their study achieved a 90.52% identification rate for the training images, and 81.58% for the test images, which they noted is similar to a human's subjective justification for the same set of images. One of the complications noted by the researchers, which likely contributed to their seemingly low rates of success in comparison to other classification studies, was the fact that the age cutoffs for varying levels of "adult" don't typically have hard divisions, but for the sake of the study, this is necessary. For example, the researchers established the cutoff between young and middle adults at 39 years old ($</= 39$ for young, $> 39$ for middle). This creates issues when individuals are right at the boundary of two age groups. To prevent similar issues with our experiment, we decided to simply classify the images as "Autistic" and "Non-Autistic" rather than trying to additionally classify the levels of autism.

In a study by Caifeng Shan, researchers used Local Binary Patterns (LBP) to describe faces. Through the application of support vector machines (SVM), they were able to achieve a 94.81% success rate in determining the gender of the subject. The main breakthrough of this study was its ability to use only real-life images in their classification. Up to this point, many of the proven studies used ideal images, most of which were frontal, occlusion-free, with a clean background, consistent lighting, and limited facial expressions. Similar to this study, our facial images are derived from real-life environments and the dataset was constructed organically.

**Classification of Autism**: A study conducted by Auman El-Baz and colleagues focused on analyzing images of cerebral white matter (CWM) in individuals with autism to determine if classification could be achieved based only on the analysis of brain images. The CWM is first segmented from the proton density MRI images, and then the CWM gyrifications are extracted and quantified. This particular approach used the cumulative distribution function of the distance map of the CWM gyrifications to distinguish between the two classes, autistic, and normal. While this study did yield successful results, the images were only taken from deceased individuals, so it's success rate classifying living individuals is still unknown. Our proposed classification system is able to achieve similar levels of accuracy (94.64%) while using significantly more subjects and only requiring an image of the individual rather than intensive, costly, brain scans and subsequent detailed analysis.

**MobileNet:** There are many different convolutional neural networks (CNN) available for image analysis. Some of the more well-known models include GoogleNet, VGG 16, Squeezenet, and AlexNet. Each of these distinct models offer different advantages, but Mobilenet has been proven to be similarly effective while greatly reducing computation time and costs. Mobilenet has shown that making their models thinner and wider has resulted in similar accuracy while greatly reducing multi-adds and parameters required for analysis (Tables 1-4).

Table 1: Depthwise Separable vs Full Convolution MobileNet[1]

| Model | ImageNet Accuracy | Million Mult-Adds | Million Parameters |
|---|---|---|---|
| Conv MobileNet | 71.7% | 4866 | 29.3 |
| MobileNet | 70.6% | 569 | 4.2 |

Table 2: Narrow vs Shallow MobileNet[1]

| Model | ImageNet Accuracy | Million Mult-Adds | Million Parameters |
|---|---|---|---|
| 0.75 MobileNet | 68.4% | 325 | 2.6 |
| Shallow MobileNet | 65.3% | 307 | 2.9 |

Table 3: MobileNet Width Multiplier[1]

| Model | ImageNet Accuracy | Million Mult-Adds | Million Parameters |
|---|---|---|---|
| 1.0 MobileNet-224 | 70.6% | 569 | 4.2 |
| 0.75 MobileNet-224 | 68.4% | 325 | 2.6 |
| 0.5 MobileNet-224 | 63.7% | 149 | 1.3 |
| 0.25 MobileNet-224 | 50.6% | 41 | 0.5 |

Table 4: MobileNet Resolution[1]

| Model | ImageNet Accuracy | Million Mult-Adds | Million Parameters |
|---|---|---|---|
| 1.0 MobileNet-224 | 70.6% | 569 | 4.2 |
| 1.0 MobileNet-192 | 69.1% | 418 | 4.2 |
| 1.0 MobileNet-160 | 67.2% | 290 | 4.2 |
| 1.0 MobileNet-128 | 64.4% | 186 | 4.2 |

In comparison with the previously mentioned models, MobileNet has shown to be just as accurate while significantly reducing the computing power necessary to run the model (Table 5). Using this knowledge, we have decided that MobileNet is a sufficient model to use for our analysis.

Table 5: MobileNet Comparison to Popular Models[2]

| Model | ImageNet Accuracy | Million Mult-Adds | Million Parameters |
|---|---|---|---|

| 1.0 MobileNet-224 | 70.6% | 569 | 4.2 |
| GoogleNet | 69.8% | 1550 | 6.8 |
| VGG 16 | 71.5% | 15300 | 138 |

## III. Methods

Our data set was obtained from Kaggle, and consists of 3,014 children's facial images total. Of these images, 1,507 of them are presumed to have autism, and the remaining 1,507 are presumed to be "healthy". Images were obtained from online, both through Facebook groups and through Google Image searches. Independent research was not conducted to determine if the individual in a picture was truly healthy or autistic. Once all the images were gathered, they were subsequently cropped so that the faces occupied the majority of the image. Prior to training, the images are split into three categories: train, validation, and test (Table 6). Images that are placed into each category must be put there manually. Therefore, repeatedly running the algorithm will generally produce the same results, assuming that the neural network ends up with the same weights. It is also worth noting that, currently, the global dataset has multiple repetitions, some of which are shared between the training, test, and validation datasets. It is therefore essential that these duplicates be cleaned out of the datasets before running the algorithm. For this case study, the duplicates have not yet been removed, which is likely improving overall accuracy.

Table 6: Dataset breakdown

| Data set | Composition | Overall data composition |
|---|---|---|
| Train | 1327 autistic 1327 healthy | 88% |
| Validation | 80 autistic 80 healthy | 5.3% |
| Test | 140 autistic 140 healthy | 9.3% |
| Total | 1507 autistic 1507 healthy | 100% |

Deep learning is broken down into three subcategories: convolutional neural networks (CNN), pretrained unsupervised networks, and recurrent and recursive networks. For this data set, we decided to use a CNN model. CNN is able to intake an image, assign importance to various objects within the image, and then differentiate objects within the image from one another. Additionally, CNNs are advantageous because the preprocessing involved is minimal compared to other methods. In this case, the input is the many images from the dataset to give an output variable: autistic or nonautistic. When looking at CNN, there are various kinds of methods to apply: LeNet, GoogLeNet, AlexNet, VGGNet, ResNet, and so forth. When trying to decide which CNN to use, it is crucial to consider what kind of data is in use, and the size of data being applied. For this instance, MobileNet is used because of the dataset: MobileNet is able to compute outputs much faster, as it can reduce computation and model size.

To perform deep learning on the dataset, MobileNet was utilized followed by two dense layers as seen in Figure 2. The first layer is dedicated to distribution, and allows customisation of weights to input into the second dense layer. Thus, the second dense layer allows for classification. The architecture of MobileNet can be reviewed in Table 7.

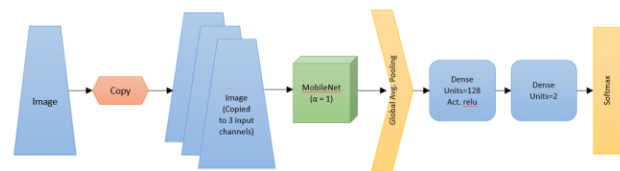

Figure 2: Algorithm architecture
Table 7: MobileNet Body Architecture

| Type / Stride | Filter Shape | Input Size |
|---|---|---|
| Conv / s2 | 3 × 3 × 3 × 32 | 224 × 224 × 3 |
| Conv dw / s1 | 3 × 3 × 32 dw | 112 × 112 × 32 |
| Conv / s1 | 1 × 1 × 32 × 64 | 112 × 112 × 32 |
| Conv dw / s2 | 3 × 3 × 64 dw | 112 × 112 × 64 |
| Conv / s1 | 1 × 1 × 64 × 128 | 56 × 56 × 64 |
| Conv dw / s1 | 3 × 3 × 128 dw | 56 × 56 × 128 |
| Conv / s1 | 1 × 1 × 128 × 128 | 56 × 56 × 128 |
| Conv dw / s2 | 3 × 3 × 128 dw | 56 × 56 × 128 |
| Conv / s1 | 1 × 1 × 128 × 256 | 28 × 28 × 128 |
| Conv dw / s1 | 3 × 3 × 256 dw | 28 × 28 × 256 |
| Conv / s1 | 1 × 1 × 256 × 256 | 28 × 28 × 256 |
| Conv dw / s2 | 3 × 3 × 256 dw | 28 × 28 × 256 |
| Conv / s1 | 1 × 1 × 256 × 512 | 14 × 14 × 256 |
| 5× Conv dw / s1 | 3 × 3 × 512 dw | 14 × 14 × 512 |
| 5× Conv / s1 | 1 × 1 × 512 × 512 | 14 × 14 × 512 |
| Conv dw / s2 | 3 × 3 × 512 dw | 14 × 14 × 512 |
| Conv / s1 | 1 × 1 × 512 × 1024 | 7 × 7 × 512 |
| Conv dw / s2 | 3 × 3 × 1024 dw | 7 × 7 × 1024 |
| Conv / s1 | 1 × 1 × 1024 × 1024 | 7 × 7 × 1024 |
| Avg Pool / s1 | Pool 7 × 7 | 7 × 7 × 1024 |
| FC / s1 | 1024 × 1000 | 1 × 1 × 1024 |
| Softmax / s1 | Classifier | 1 × 1 × 1000 |

For our MobileNet, an alpha of 1 and depth multiplier of 1 were utilized, thus we use the most baseline version of MobileNet. In order to make binary predictions from MobileNet, two fully-connected layers are appended to the end of the model. The first is a dense layer with 128 neurons (L2 regularization = 0.015, ReLu activation) which is then connected to the prediction layer which only has two outputs (softmax activation). A dropout of 0.4 is applied to the first layer to prevent overfitting. The final output is a binary classification of either "autistic" or "non-autistic."

The algorithm was run on an ASUS laptop with an Intel Core i7-6700HQ CPU at 2.60 GHz and 12 GB of RAM. Data was broken into batch sizes of 80. Upon completion of training and initial testing, the user can request additional training epochs.

## IV.    Results

Training completed after ~15 epochs, yielding a test accuracy of 94.64%. Figure 1 shows how the loss of the training and test set changed with the continual addition of one epoch at a time. Figure 2 shows how accuracy changed for training, validation, and test with the continual addition of one epoch.

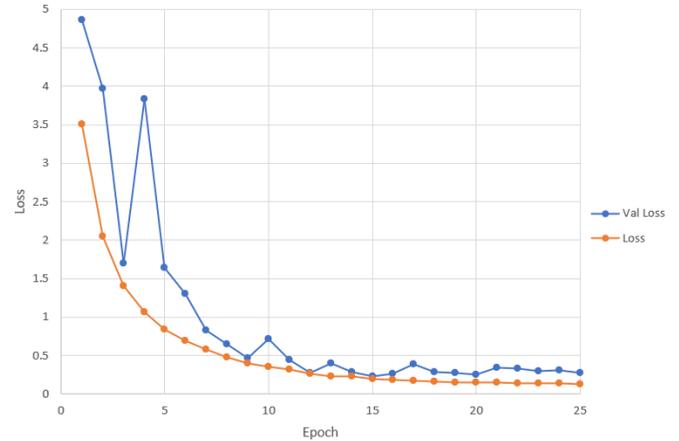

Figure 1: Model loss

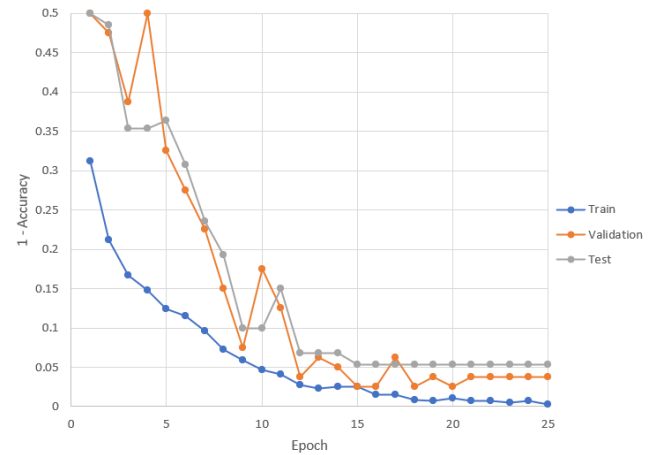

Figure 2: Model accuracy

During training, the weights that gave the validation set the highest accuracy were always stored. Therefore, if the accuracy decreased during a training set, there would be no ultimate loss of accuracy on the test set (Figure 1). Similarly, if accuracy on the validation set decreased, the learning rate would also decrease during the next training session. Each epoch required approximately 10 minutes to run.

These preliminary results are very promising. Currently, there are many issues with the dataset that was used including duplicate images, improper age ranges, and lack of validation about the conditions of the individuals in each photo. Improving the data set could result in better results.

## V.    Conclusion

In conclusion, while the statistics on how many children are diagnosed with autism are somewhat low, it is extremely important to diagnose as early as possible in order to provide the correct care for the patient. Additionally, the statistics on diagnosed children may be low because the method to accurately diagnose a child is somewhat ineffective. Thus, our classifier could prove to be very useful in diagnosing more children accurately. Our results show we successfully achieved a high accuracy of 94.6%, meaning that it was able to identify a child with or without autism correctly. In order to improve accuracy, cleaning the dataset would certainly help. Duplicates may falsely increase our test accuracy if an image is also in the training category. With more information about the individuals in the pictures, we could also ensure that age distributions are similar between the two populations. Currently, autism is rarely diagnosed in young children, so we would also ensure that no young pictures are in our dataset. Similarly, we could ensure that each category is "pure," preventing false-positives and false-negatives. With these improvements, we would hope to get our accuracy to about 95%.

The success of this algorithm may also imply that other diseases can be diagnosed using only a picture, saving valuable time and resources in regards to diagnosing other diseases and conditions. Down's Syndrome, for example, is another disease that markedly alters the facial features of those it afflicts. It is possible that, given sufficient and good data, our algorithm could distinguish between individuals with the disease and individuals that do not have it.